\documentclass[letterpaper,journal]{IEEEtran}
\usepackage{amsmath,amsfonts}
\usepackage{algorithmic}
\usepackage{algorithm}
\usepackage{array}
\usepackage[caption=false,font=normalsize,labelfont=sf,textfont=sf]{subfig}
\usepackage{textcomp}
\usepackage{stfloats}
\usepackage{url}
\usepackage{verbatim}
\usepackage{graphicx}
\usepackage{cite}
\usepackage{amsmath}
\usepackage{adjustbox}
\usepackage{booktabs}
\usepackage{multirow}
\usepackage{color} 
\begin{document}

\title{Towards Controllable Low-Light Image Enhancement: A Continuous Multi-illumination Dataset and Efficient State Space Framework}

\author{Hongru Han, 
        Tingrui Guo, 
        Liming Zhang,~\IEEEmembership{Senior Member,~IEEE,}
        Yan Su, 
        Qiwen Xu,
        and Zhuohua Ye
\thanks{Hongru Han, Liming Zhang, Qiwen Xu, and Zhuohua Ye are with the Department of Computer and Information Science, University of Macau, Taipa, Macau 999078, China (e-mail: yc37462@connect.um.edu.mo; lmzhang@um.edu.mo; qwxu@um.edu.mo; yc37977@connect.um.edu.mo).}%
\thanks{Tingrui Guo is with  the School of Automation, China University
of Geosciences, Wuhan 430074, China
(e-mail: guotingrui@cug.edu.cn).}%
\thanks{Yan Su is with the Department of Electromechanical Engineering, University of Macau, Taipa, Macau 999078, China (e-mail: yansu@um.edu.mo).}%
\thanks{Corresponding author: Liming Zhang.}
}

\markboth{Journal of \LaTeX\ Class Files,~Vol.~14, No.~8, August~2021}%
{Shell \MakeLowercase{\textit{et al.}}: A Sample Article Using IEEEtran.cls for IEEE Journals}


\maketitle

\begin{abstract}
Low-light image enhancement (LLIE) has traditionally been formulated as a deterministic mapping. However, this paradigm often struggles to account for the ill-posed nature of the task, where unknown ambient conditions and sensor parameters create a multimodal solution space. Consequently, state-of-the-art methods frequently encounter luminance discrepancies between predictions and labels, often necessitating "gt-mean" post-processing to align output luminance for evaluation. To address this fundamental limitation, we propose a transition toward Controllable Low-light Enhancement (CLE), explicitly reformulating the task as a well-posed conditional problem. To this end, we introduce CLE-RWKV, a holistic framework supported by Light100, a new benchmark featuring continuous real-world illumination transitions. To resolve the conflict between luminance control and chromatic fidelity, a noise-decoupled supervision strategy in the HVI color space is employed, effectively separating illumination modulation from texture restoration. Architecturally, to adapt efficient State Space Models (SSMs) for dense prediction, we leverage a Space-to-Depth (S2D) strategy. By folding spatial neighborhoods into channel dimensions, this design allows the model to recover local inductive biases and effectively bridge the "scanning gap" inherent in flattened visual sequences without sacrificing linear complexity. Experiments across seven benchmarks demonstrate that our approach achieves competitive performance and robust controllability, providing a real-world multi-illumination alternative that significantly reduces the reliance on gt-mean post-processing.
\end{abstract}

\begin{IEEEkeywords}
Low-light image enhancement, controllable low-light enhancement, Receptance Weighted Key Value, HVI Color Space.
\end{IEEEkeywords}

\section{Introduction}
\IEEEPARstart{L}{ow-light} image enhancement (LLIE) aims to restore visibility and detail from inputs degraded by insufficient illumination. With the advent of deep learning, numerous methods~\cite{SID, KinD, Retinexformer, MambaLLIE, SGDT, SGF, BiFormer} have surpassed traditional priors~\cite{he1} by learning a deterministic mapping from low-light to normal-light images.

\begin{figure}[t]
  \centering
  \includegraphics[width=\columnwidth]{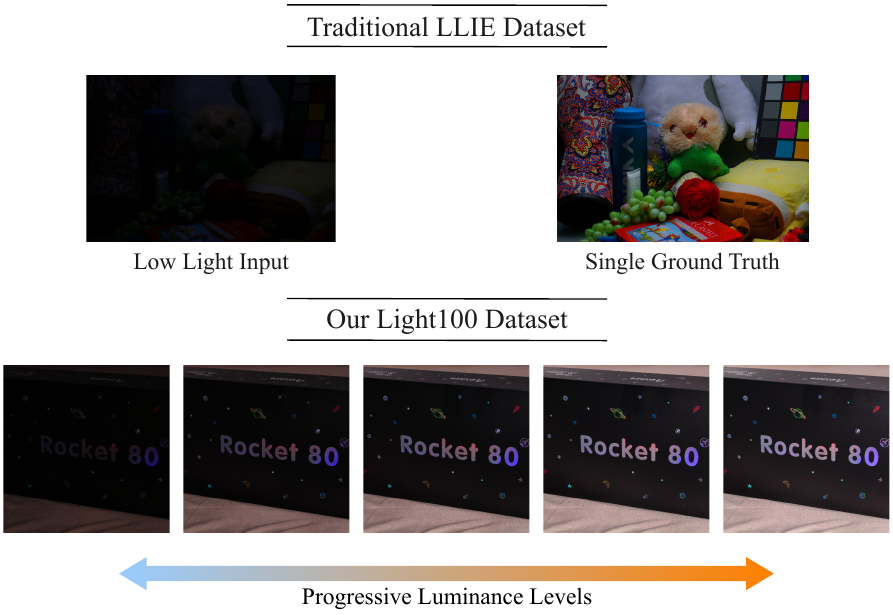} 
  \caption{Comparison between static ``one-to-one'' benchmarks (top) and our Light100 dataset (bottom). While traditional datasets are limited to a single ground-truth mapping, Light100 captures 100 progressive lighting levels. This allows for a flexible mapping governed by the target luminance level $\beta$, accommodating various user preferences and ambient conditions.}
  \label{fig:lolv1}
\end{figure}

Despite these advances, the prevailing “one-to-one” approach is found to struggle with preserving details while adapting to diverse illumination conditions. First, LLIE is fundamentally an ill-posed problem, since the observed low-light image is determined by both scene reflectance and latent illumination parameters. Consequently, a single input can map to a manifold of valid normal-light representations depending on ambient conditions and sensor settings. As shown in Fig.~\ref{fig:lolv1}, most benchmarks collapse this ambiguity into a single reference image, inevitably creating a luminance gap between model outputs and the provided target. This discrepancy can introduce a subtle optimization bias: since pixel-wise losses react strongly to overall brightness changes, models often favor matching luminance over correcting subtle defects such as noise or color inaccuracies.

To bridge this shortcoming, existing methods~\cite{LLFlow, PyDiff, LLFormer, GSAD, CIDNet} commonly apply a post-processing fix which is matching output brightness to the ground-truth mean (gt-mean) so that texture and structure can be assessed independently. While specialized objectives~\cite{gtmeanloss} attempt to alleviate this issue during training, they primarily address the symptoms rather than the underlying architectural limitation: the attempt to force a single, fixed output onto a problem that naturally admits many valid solutions. Second, the existing approaches often overlook the subjective nature of human perception. In many practical scenarios, users may prefer to retain a specific ``nighttime'' atmosphere rather than an automated transition to daylight-like appearance. Such controllability is becoming essential in applications ranging from customized smartphone photography to creative workflows that involve human guidance.

Controllable Low-light Enhancement (CLE) redefines low-light restoration as a conditional generation task, using target luminance as input. This transforms the inherently ill-posed inverse problem into a well-posed one. However, systematically establishing CLE faces challenges spanning data and architecture that go beyond merely extending traditional LLIE frameworks.

The first challenge is the lack of a standardized, real-world multi-illumination benchmark for controllable enhancement. Existing works~\cite{ReCoRo, CLE-Diff} are often constrained by the ``one-to-one'' nature of standard LLIE datasets, necessitating the use of synthetic luminance adjustments to simulate controllability. However, such synthetic methods fail to capture the complex, non-linear signal degradation inherent in true real-world illumination changes. Crucially, real multi-illumination data exposes a hidden pitfall: dim reference images embed both the desired brightness level and the degradations of low-light capture, forcing models to reproduce noise and color bias along with luminance.

The second challenge involves the trade-off between efficiency and capability. Since CLE requires real-time feedback, the heavy compute costs of Transformers~\cite{Retinexformer} or the iterative sampling of Diffusion models~\cite{CLE-Diff} are often prohibitive. Alternatively, while State Space Models (SSMs)~\cite{mamba, rwkv} provide linear complexity, their visual adaptations~\cite{VMamba, VRWKV} still suffer from spatial disruption and information decay. Flattening 2D images severs vertical proximity via a 'scanning gap,' forcing models to recover local dependencies over long 1D distances. Furthermore, recurrent state updates often lose fine-grained details, compromising the structural fidelity essential for restoration.

A holistic framework is introduced to systematically address these challenges, demonstrating that incorporating brightness-aware conditioning provides a more effective alternative to conventional LLIE methods. To overcome data limitations, Light100 is proposed as the first real-world multi-illumination benchmark, comprising 100 lighting conditions that vary smoothly across a continuous illumination spectrum. Crucially, by leveraging the HVI color space~\cite{CIDNet}, a novel Noise-Decoupled Supervision Strategy is proposed. It generates hybrid reference images that preserve target brightness while incorporating clean textures from well-lit scenes, enabling the model to learn illumination adjustment without absorbing sensor noise. Architecturally, CLE-RWKV is proposed, which incorporates a Space-to-Depth (S2D) design to reconcile long-range dependency modeling with the local inductive biases essential for image restoration. By combining Local-RWKV (L-RWKV) blocks with a parameter-free periodic shuffle strategy, this framework folds spatial neighborhoods into channel dimensions, effectively recovering high-frequency details while mitigating long-range forgetting in flattened visual sequences.

The main contributions of this work are summarized as follows:
\begin{itemize}
\item We theoretically justify the transition from LLIE to a controllable enhancement pipeline, showing that conditioning on target luminance transforms the ill-posed problem into a well-posed one, thus avoiding the need for post-processing adjustments like gt-mean.

\item Light100, the first dataset of real-world images captured under 100 calibrated lighting levels is introduced. Meanwhile, a noise-decoupled supervision strategy is introduced to decouple  illumination from color and textures.

\item CLE-RWKV, a specialized adaptation of State Space Models optimized for dense and controllable restoration, is introduced. By integrating Local-RWKV (L-RWKV) blocks within a Space-to-Depth (S2D) mechanism, the proposed architecture effectively bridges the scanning gap inherent in 1D sequences and recovers local spatial consistency. This design achieves a superior balance between real-time interactivity and computationally efficient high-fidelity restoration.
\end{itemize}

\section{Related Work}
\subsection{From LLIE to CLE: Architectures and Paradigms}
The methodology of low-light enhancement has transitioned from local CNN-based restoration~\cite{RetinexNet, KinD} to global modeling via Vision Transformers~\cite{Retinexformer, LLFormer} and efficient State Space Models (SSMs)~\cite{MambaLLIE, URWKV}. Despite these advances, the majority of existing methods are confined to a fixed-target paradigm, mapping a single input to a pre-defined ground truth. This rigid formulation fails to accommodate varying ambient conditions or user-specific preferences.

To address this, recent studies have explored Controllable Low-light Enhancement (CLE) by introducing explicit conditioning. However, existing CLE approaches face limitations in either real-world multi-illumination modeling or efficiency. For instance, Xu et al.~\cite{ReCoRo} employs a control scalar primarily as a heuristic interpolation weight between the low-light input and the normal-light reference, rather than representing a calibrated real-world luminance level. On the other hand, Yin et al.~\cite{CLE-Diff} leverages generative diffusion priors to achieve more precise manipulation, but its iterative sampling process incurs prohibitive computational overhead and high latency, hindering real-time user interaction. Therefore, achieving accurate luminance control aligned with real-world transitions without sacrificing the real-time throughput necessary for interactive systems remains a critical challenge.

\subsection{Benchmarks for LLIE}
The progression of LLIE research has relied heavily on the availability of large-scale paired datasets. Pioneering benchmarks such as SID~\cite{SID} and LOL~\cite{RetinexNet} enabled end-to-end training in the raw and sRGB domains, respectively.
Later works introduced more complex variables; for instance, LOL-v2~\cite{LOLv2} emphasized scene diversity, while SMID~\cite{SMID} provided a vast collection of high-noise RAW image pairs. SDSD~\cite{SDSD} further shifted the focus toward the video domain. However, these benchmarks remain limited by a deterministic, fixed-target paradigm. Whether employing a one-to-one~\cite{RetinexNet, LOLv2} or many-to-one~\cite{SID, SMID} strategy, they invariably map degraded inputs to a single, pre-defined reference. Such a static mapping cannot accommodate the demand for controllable results that allow users to select their desired level of illumination. While some recent attempts have explored controllable settings, the field still lacks a rigorous benchmark that provides paired, multi-illumination ground truths to systematically quantify the precision of controllable enhancement.

\subsection{Degradation Decoupling Strategies}
Low-light images in the sRGB domain suffer from deeply coupled degradations, where diminished illumination is inseparably linked with noise amplification and color distortion. Naively enhancing luminance often exacerbates these artifacts. Consequently, decoupling these components has become a standard strategy for robust enhancement. The most established paradigm is rooted in Retinex theory~\cite{RetinexNet, KinD, Retinexformer, Diff-retinex, MS-Retinex}, which explicitly separates the image into reflectance (physical property) and illumination (variable property) components. Beyond Retinex, other strategies have explored alternative domains. Frequency-domain methods~\cite{FourierDiff, FourLLIE, WalMaFa, PPLIE} utilize Fourier or Wavelet transforms to separate low-frequency illumination from high-frequency noise and texture, allowing for targeted restoration. More recently, color-space decoupling has gained traction. Yan et al.~\cite{CIDNet} introduced the HVI color space, which mathematically decouples intensity from chromaticity. The HVI color space is leveraged to construct the supervision strategy in this work. Unlike Retinex approaches that rely on estimation, HVI offers an explicit, training-free decomposition. This property is crucial for our framework, as it facilitates the seamless generation of noise-decoupled targets, allowing us to strictly isolate illumination control from texture restoration without introducing ambiguity from auxiliary decomposition models.

\section{Proposed Method}
\label{sec:method}

\subsection{Theoretical Analysis: The Deterministic Barrier}
\label{sec:theory}

\begin{figure}[t]
  \centering
  \includegraphics[width=1.0\linewidth]{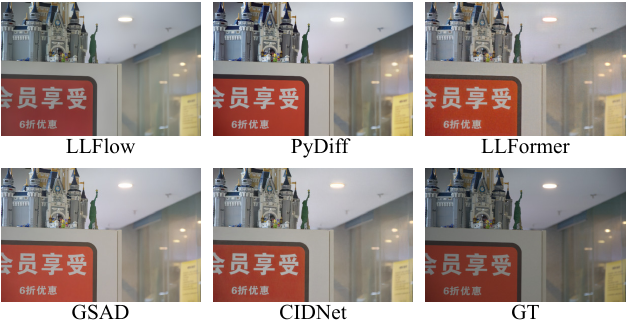}
  \caption{Visual comparison of state-of-the-art deterministic enhancement methods on the LOL-v2-Real dataset. Most advanced models produce outputs with a higher average luminance than the Test Ground Truth (GT), demonstrating a systematic luminance mismatch that stems from the deterministic mapping paradigm.}
  \label{fig:visual_bias}
\end{figure}

\begin{figure}[t]
  \centering
  \includegraphics[width=1.0\linewidth]{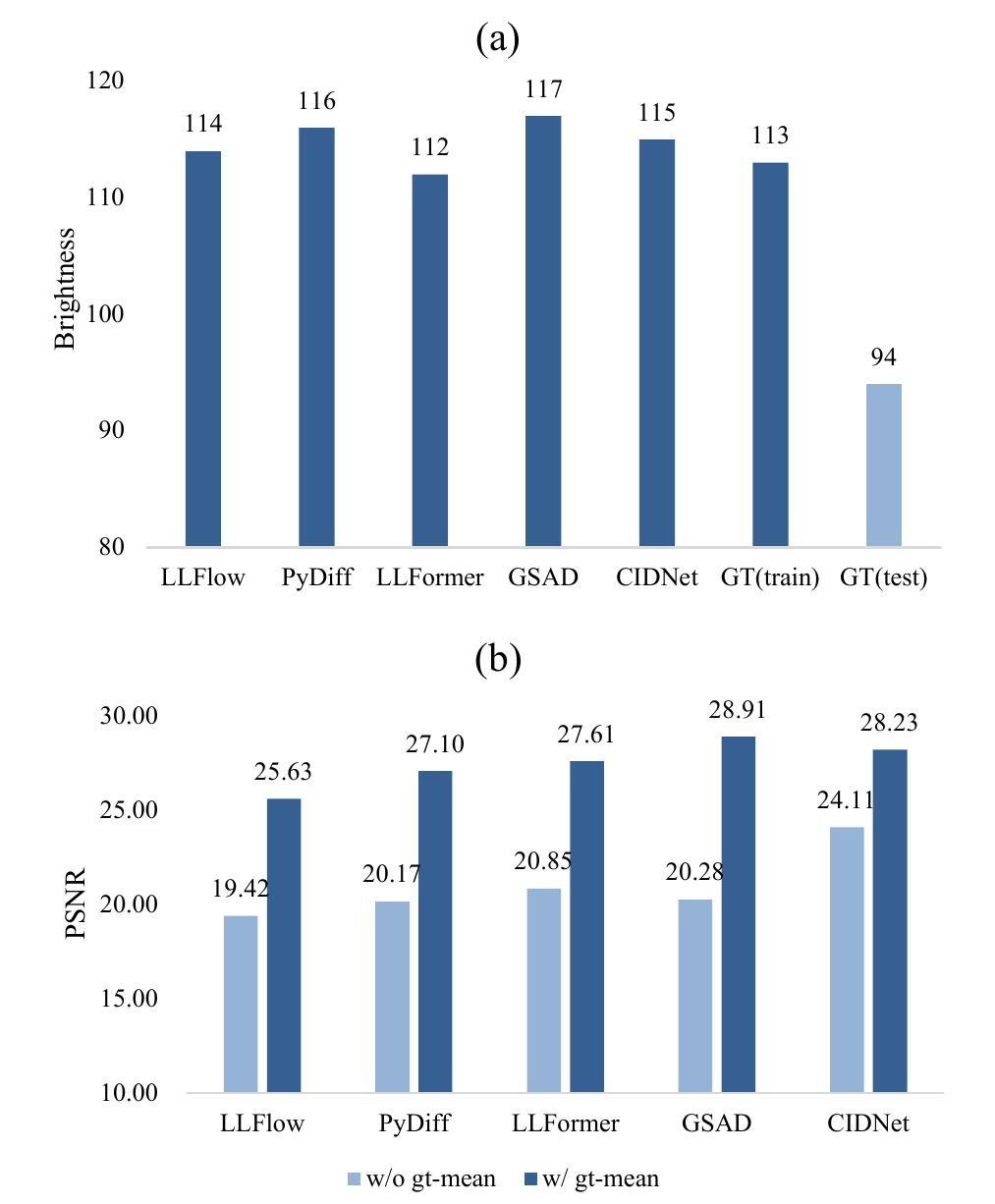}
  \caption{Quantitative analysis of evaluation bias on the LOL-v2-Real dataset. (a) Luminance distribution analysis: outputs of SOTA methods tend to align with the global Training GT mean rather than adapting to the specific illumination of individual test samples. (b) The "gt-mean" heuristic is commonly employed to evaluate the fidelity of texture and color by neutralizing the global luminance gap, leading to substantial PSNR gains. Here, w/o and w/ denote evaluations without and with this post-hoc adjustment, respectively.}
  \label{fig:bias_analysis}
\end{figure}

The task is typically treated by recent methods as a deterministic regression mapping low-light inputs $x$ to normal-light targets $y$. However, empirical results on the LOL-v2-Real dataset reveal a consistent systematic bias. As shown in Fig.~\ref{fig:visual_bias}, existing methods such as~\cite{LLFlow, PyDiff, LLFormer, GSAD, CIDNet} generate outputs that are consistently brighter than the ground truth image. Specifically, these models tend to disregard the unique illumination conditions of individual test scenes and instead constrain the output brightness to align with the average luminance observed in the training set. This bias arises from the standard training process, which encourages the network to predict a safe average luminance that minimizes the overall reconstruction error. Owing to an excessive focus on matching this global brightness level, the models frequently fail to recover fine textures or preserve natural color fidelity.

\subsubsection{Statistical Bound: The Regression-to-Median Dilemma}
LLIE is an ill-posed inverse problem. A single low-light observation $x$ may correspond to multiple plausible normal-light reconstructions $y$, depending on the  unobserved ambient illumination conditions. This indicates that the conditional distribution $P(y|x)$ exhibits multiple peaks rather than converging to a single deterministic outcome.

In standard deep learning~\cite{BishopPRML}, a deterministic model $\mathcal{F}$ trained with pixel-wise $\ell_1$ reconstruction loss aims to minimize the expected mean absolute error per pixel:
\begin{equation}
\mathcal{F}^* = \arg\min_{\mathcal{F}} \mathbb{E}_{x,y} [\| y - \mathcal{F}(x) \|_1].
\end{equation}
The mathematical solution to this objective corresponds to the conditional median of the target distribution:
\begin{equation}
\mathcal{F}^*(x) = \text{median}_{P(y|x)}[y].
\end{equation}

However, as analyzed in~\cite{Mathieu2016}, while the $\ell_1$ loss is more robust to outliers than $\ell_2$, it still produces blurry or visually inconsistent predictions in multimodal scenarios. This happens because the pixel-wise median, when computed across multiple plausible lighting conditions, often yields an intermediate value that lacks the characteristics of any specific condition. Consequently, the generated image appears unnatural and fails to conform to the statistical distribution of real photographs. Since the actual illumination (e.g., light intensity or exposure) is hidden during inference, deterministic models treat $P(y|x)$ as a mixture of various lighting conditions from the training set. To minimize the overall $\ell_1$ risk, these networks learn to predict a static median mapping based on training statistics.

Fig.~\ref{fig:bias_analysis}(a) empirically validates this bias. Empirical results indicate that outputs from existing deterministic methods cluster around the median brightness of the training set, regardless of the specific requirements of the test target. This suggests that model behavior is dominated by training-set priors, rather than adapting to the unique expected illumination of each test scene.

The CLE paradigm addresses this limitation by treating $\beta$ as a known condition. This reformulates the problem from estimating the median of a multi-peak distribution to identifying a single target component, thereby substantially reducing the conditional entropy:
\begin{equation}
H(y|x, \beta) \ll H(y|x).
\end{equation}
By incorporating the physical prior $\beta$, the uncertainty of multiple peaks is collapsed into a single, well-defined state, enabling the model to adaptively reach the correct luminance without falling into the regression-to-median trap.

\subsubsection{Physical Limitations of the gt-mean Heuristic}
The gt-mean heuristic is commonly used to assess texture and color fidelity in deterministic models by removing global luminance discrepancies. This is achieved by linearly scaling the model output to align with the target’s mean luminance, thereby illustrating the model’s hypothetical performance if luminance bias were absent, as illustrated in Fig.~\ref{fig:bias_analysis}(b). However, this post adjustment is fundamentally inaccurate from a physical perspective and fails to represent true restoration performance for two key reasons.

\paragraph{Noise Variance Amplification}
A low-light observation can be described as $x = S + n$, where $S$ is the signal and $n$ is noise. When the output is linearly amplified by a gain factor $k>1$ to match the target brightness, both the signal and noise are scaled accordingly $y_{scale} = kS + kn$. This causes the noise variance to increase quadratically $\text{Var}(y_{scale}) = k^2 \cdot \text{Var}(n)$, frequently introducing visible noise in the restored image. In contrast, the proposed CLE-RWKV learns a non-linear mapping $\mathcal{F}(x, \beta)$ that integrates denoising with enhancement, thereby producing bright outputs without amplifying noise linearly.

\paragraph{Non-linear ISP Distortion}
The primary limitation of the gt-mean heuristic is that it ignores the non-linearity of the Image Signal Processing (ISP) pipeline. Real-world cameras employ an ISP that applies a non-linear transformation $\mathcal{T}$ encompassing gamma correction and tone mapping to map raw radiance $R$ to sRGB pixels. Real-world enhancement happens in the radiance domain: $y_{true} = \mathcal{T}(k \cdot R)$. However, the gt-mean heuristic scales the processed pixels: $y_{heuristic} = k \cdot \mathcal{T}(R)$. Because $\mathcal{T}$ is highly non-linear, these two operations do not commute:
\begin{equation}
\| \mathcal{T}(k \cdot R) - k \cdot \mathcal{T}(R) \| > 0.
\end{equation}
This gap means that linear scaling cannot correctly replicate the contrast and color of natural images (see Sec. \ref{sec:Exp} for visual evidence). The proposed CLE framework, acting as a non-linear approximator, implicitly learns to reverse the ISP steps and apply enhancement correctly: $\mathcal{F}(x, \beta) \approx \mathcal{T}(k \cdot \mathcal{T}^{-1}(x))$, maintaining physical fidelity across all luminance levels.

\subsection{The Light100 Benchmark: Enabling Continuous Control}
\label{sec:physical_basis}

Conventional datasets are generally limited to binary ``low'' and ``high'' exposure pairs, restricting controllable models to interpolation between extreme states. Realizing a truly continuous control mapping $\mathcal{F}(x, \beta)$ requires a dataset that densely samples the illumination space.

The Light100 benchmark is established as a real-world multi-illumination dataset with physically-calibrated references, captured under rigorously controlled hardware conditions. In contrast to existing datasets, 100 progressive lighting levels were recorded for each of the 275 scenes by systematically modulating the light source power from 1W to 100W. To ensure that image variations are driven solely by illumination levels, all camera parameters were fixed for each specific scene, including ISO 72, Aperture $f/2.6$, and an exposure time of 1/50s or 1/100s. To ensure the integrity of the data manifold, the exposure time for each scene was calibrated based on the maximum illumination level (100W). This protocol explicitly prevents highlight saturation (clipping) even in scenes with highly reflective objects, ensuring that the pixel intensity remains within the sensor's linear response range across all 100 progressive levels.

To facilitate the learning of real-world process, the control signal $\beta \in [0, 1]$ is defined as the normalized mean luminance associated with each illumination level. Although the camera's internal processing is non-linear, the mapping from light power to pixel intensity remains strictly monotonic. With 100 densely sampled levels, $\beta$ serves as a continuous coordinate that enables the learning of the fine-grained mapping:
\begin{equation}
\mathcal{M}_{phy} = \{ (x_i, \beta_i) \mid \beta_i = \text{Mean}(y_i), \ i = 1, \dots, 100 \}.
\end{equation}
Crucially, this continuity enables both the smooth learning of the illumination embedding and the precise evaluation of controllability across the full range of illumination, bridging the gap between real-world illumination levels and digital pixel values.

\begin{figure*}[t]
  \centering
  \includegraphics[width=1.0\linewidth]{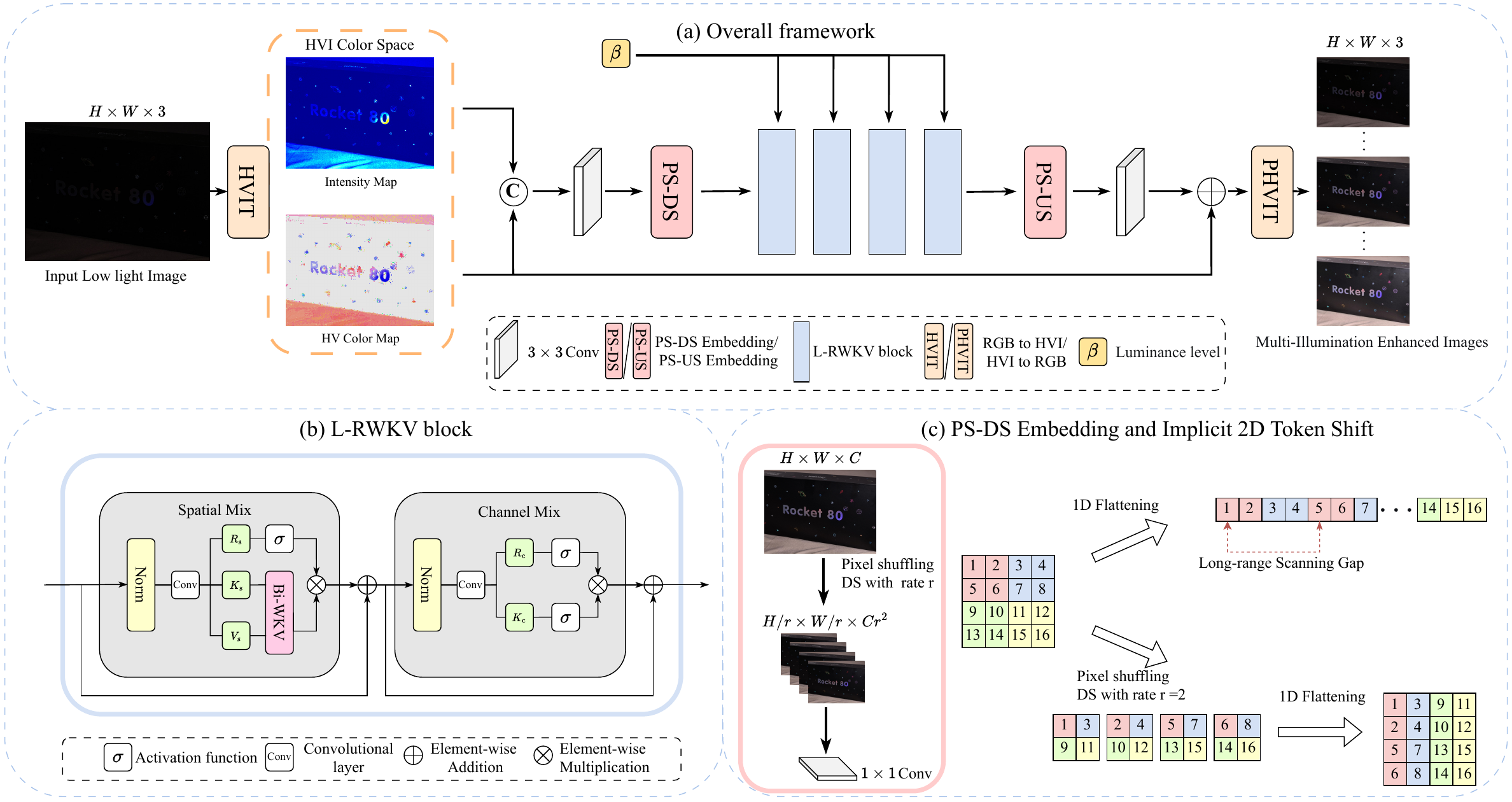}
  \caption{Overview of the Proposed CLE-RWKV Framework. (a) The overall pipeline operates in the decoupled HVI color space, where target luminance commands ($\beta$) modulate features via FiLM. (b) The S2D-Net Backbone employs a split-transform-merge strategy, where the L-RWKV Block serves as the core processor. (c) Analysis of the Periodic Shuffle Strategy. Left: The PS-DS Embedding layer (Pixel Shuffle Downsampling with $r$ followed by $1\times 1$ Conv) demonstrates that semantic topology and features are preserved even at lower resolutions. Right: Visualization of the implicit 2D Token Shift. Compared to conventional 1D flattening, our periodic shuffle strategy folds spatial neighborhoods into the channel dimension, effectively bridging the ``scanning gap'' and enabling the L-RWKV block to capture local details while mitigating long-range forgetting.}
  \label{fig:framework}
\end{figure*}

\subsection{CLE-RWKV Architecture}
As illustrated in Fig.~\ref{fig:framework}(a), the proposed CLE-RWKV framework operates in three stages: HVI-based decoupling, illumination-conditioned feature extraction, and reconstruction via a perceptual-inverse transform.

\subsubsection{Illumination-Aware Feature Modulation}
The input sRGB image is first transformed into the HVI domain to separate the texture-rich color maps ($\mathrm{I}_\mathrm{HV}$) from the intensity map ($\mathrm{I}_\mathrm{I}$). These are concatenated and projected into a shallow feature space. User control is incorporated by encoding the scalar target luminance $\beta$ into a high dimensional illumination embedding via learnable interpolation~\cite{CLE-Diff}. This embedding is fed into a Multi-Layer Perceptron (MLP) to generate affine parameters $(\gamma, \mu)$, which modulate the intermediate features via Feature-wise Linear Modulation (FiLM) layers. This mechanism, shown in the conditional branch of Fig.~\ref{fig:framework}(a), ensures that the restoration process is strictly conditioned on the target exposure level, effectively modeling $P(y|x, \beta)$.

\subsubsection{Space-to-Depth (S2D) Design}
The backbone of CLE-RWKV adopts a Space-to-Depth (S2D) design to adapt State Space Models (SSMs) for dense prediction. While standard RWKV~\cite{rwkv} offers linear complexity, flattening 2D feature maps severs vertical continuity. This creates a ``scanning gap'' where vertically adjacent pixels become distant in the 1D sequence, hindering the modeling of local structures.

\paragraph{PS-DS Embedding}
To resolve this efficiently, the PS-DS Embedding is employed. As illustrated in Fig.~\ref{fig:framework}(c), the Pixel Unshuffle operation~\cite{shi2016real, ibrahem2025pixel} is implemented to rearrange inputs $X \in \mathbb{R}^{H \times W \times C}$ into a shape of $\frac{H}{r} \times \frac{W}{r} \times (C r^2)$ (default $r=4$), followed by a $1 \times 1$ convolution. 

This design is proved pivotal for two reasons. First, as visualized in the left part of Fig.~\ref{fig:framework}(c), the 2D spatial topology is preserved even at lower resolutions. Second, and crucially, the local $r \times r$ neighborhood is folded into the channel dimension of a single token. Consequently, the subsequent linear projections are enabled to act as implicit spatial convolutions. This mimics a Token Shift operation in 2D space, as neighboring pixels (now channels) can be essentially ``seen'' via channel mixing, effectively bridging the scanning gap. Furthermore, by compacting the sequence length by a factor of $r^2$, the long-range forgetting typically found in flattened sequences is significantly mitigated.

\paragraph{L-RWKV Block}
As the core processor (Fig.~\ref{fig:framework}(b)), the L-RWKV block treats these compacted tokens as spatial patches to recover local inductive bias. It comprises two sub-modules:

\textbf{Spatial Mix:} This module aggregates global context. Adopting the design from~\cite{restorerwkv}, a depth-wise convolution is employed for local aggregation. The normalized input is processed by this convolution and subsequently projected by linear weights $W_i$ to yield Receptance $R_s$, Key $K_s$, and Value $V_s$:
\begin{equation}
    \{R_{\mathrm{s}}, K_{\mathrm{s}}, V_{\mathrm{s}}\} = \{ \mathrm{DWConv}_{i}(\mathrm{LN}(X)) W_{i} \}_{i \in \{R, K, V\}}.
\end{equation}
A bidirectional WKV (Bi-WKV)~\cite{VRWKV} mechanism is then utilized to aggregate information: $w k v_{t}=\operatorname{Bi-WKV}\left(K_{s}, V_{s}\right)_{t}$. The final output is generated via a gating mechanism: $O_{\mathrm{s}} = (\sigma(R_{\mathrm{s}}) \odot wkv) W_{O}$.

\textbf{Channel Mix:} Handling channel-wise fusion, this module generates $R_c$ and $K_c$:
\begin{equation}
    \{R_{\text{c}}, K_{\text{c}}\} = \{ \mathrm{DWConv}_{i}(\mathrm{LN}(X)) W_{i} \}_{i \in \{R, K\}}.
\end{equation}
$V_c$ is derived from $K_c$ via a squared ReLU, and the output is gated by $O_{\text{c}}=\left(\sigma\left(R_{\text{c}}\right) \odot V_{\text{c}}\right) W_{O}$. Finally, the spatial resolution is restored by a PS-US Embedding layer.

\begin{figure}[t]
  \centering
  \includegraphics[width=1.0\linewidth]{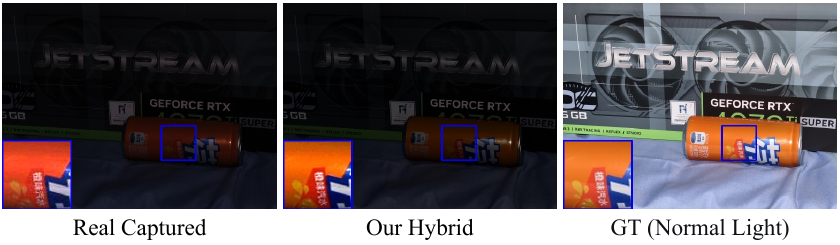}
    \caption{Resolving the Intensity-Chroma Conflict. (Left) Real physical captures at low illumination suffer from diminished SNR, manifesting as both stochastic noise and severe color shifts. (Middle) The hybrid target synthesizes the target-level intensity ($\mathbf{I}_{max}$) with the pristine chromaticity and texture ($\hat{\mathbf{H}}, \hat{\mathbf{V}}$) extracted from the high-quality reference (Right). This strategy provides a noise-free, color-accurate guidance that decouples illumination control from intrinsic degradation restoration.}
    \label{fig:hvi_paradox}
\end{figure}

\subsection{Noise-Decoupled Supervision}
\label{sec:optimization}

While real-world continuity is provided by the Light100 benchmark, direct training on raw captures remains challenging. Images captured under extremely low illumination exhibit a low Signal-to-Noise Ratio (SNR) and pronounced color casts due to sensor limitations. Using such degraded images as supervision targets forces the network to overfit to sensor-specific noise patterns and reproduce systematic color shifts, rather than learning pure illumination enhancement, as illustrated in Fig.~\ref{fig:hvi_paradox} (Left).

To address this issue, a Noise-Decoupled Supervision Strategy is proposed. The HVI color space~\cite{CIDNet} is employed to decouple global illumination intensity from noise-sensitive chromatic components. As shown in Fig.~\ref{fig:hvi_paradox} Right, a hybrid ground truth is constructed by combining the target-level intensity with the clean chromaticity and texture from a well-exposed reference image. This hybrid signal preserves real-world accurate luminance while ensuring perceptual cleanliness in color and texture.

\subsubsection{Hybrid Target Synthesis}
The goal is to synthesize a target image with the desired brightness level $\beta$, while preserving the high-fidelity chromatic and textural quality of an optimal reference image. Given an input $x$ and its corresponding raw capture $\mathrm{I}^{\beta}$ at the target illumination level $\beta$, the HVI Transformation (HVIT) is utilized to decompose each into its constituent components:
\begin{equation}
\begin{aligned}
(\hat{\mathbf{H}}^{ref}, \hat{\mathbf{V}}^{ref}, \_) &= \text{HVIT}(\mathrm{I}^{ref}), \\
(\_, \_, \mathbf{I}_{max}^{\beta}) &= \text{HVIT}(\mathrm{I}^{\beta}).
\end{aligned}
\end{equation}
Here, $\mathbf{I}_{max}^{\beta}$ provides the exact physical intensity for condition $\beta$, while $(\hat{\mathbf{H}}^{ref}, \hat{\mathbf{V}}^{ref})$ provides the clean texture and color. The final hybrid target is reconstructed via the Perceptual-inverse HVI Transformation (PHVIT):
\begin{equation}
\hat{\mathrm{I}}_{\text{Hybrid}}^\beta = \text{PHVIT}(\hat{\mathbf{H}}^{ref}, \hat{\mathbf{V}}^{ref}, \mathbf{I}_{max}^{\beta}).
\end{equation}
This strategy acts as a selective filter, guiding the model to focus on luminance modulation without being penalized for suppressing the shot noise inherent in the original dim targets.

\subsubsection{Objective Function}
The objective function is defined directly against these synthesized targets $\hat{\mathrm{I}}_{\text{Hybrid}}^\beta$. A composite loss $\mathcal{L}_\text{total}$ is formulated to constrain both the sRGB output and the intermediate HVI components:
\begin{equation}
\mathcal{L}_\text{total} = \mathcal{L}_\text{base}(\hat{\mathrm{I}}, \hat{\mathrm{I}}_{\text{Hybrid}}^\beta) + \lambda \mathcal{L}_\text{base}(\hat{\mathrm{I}}_\mathrm{HVI}, \hat{\mathrm{I}}_\mathrm{Hybrid-HVI}),
\label{eq:total_loss}
\end{equation}
where $\lambda$ balances the two domains. The base loss $\mathcal{L}_\text{base}$ handles pixel-level accuracy and perceptual quality:
\begin{equation}
\begin{split}
    \mathcal{L}_\text{base} = w_{\ell_1} \mathcal{L}_{\ell_1} + w_\text{SSIM} \mathcal{L}_\text{SSIM} + w_\text{Edge} \mathcal{L}_\text{Edge} + w_\text{LPIPS} \mathcal{L}_\text{LPIPS}.
\end{split}
\label{eq:base_loss}
\end{equation}

Following CIDNet~\cite{CIDNet}, the loss weights are configured to ensure that the model converges to a solution that is accurately aligned with target real-world luminance levels while remaining perceptually clean in texture.

\section{Experiment}
\label{sec:Exp}

\subsection{Datasets and Implementation Details}

\subsubsection{Datasets} 
To comprehensively evaluate our framework, a combination of standard fixed-target benchmarks and our proposed controllable benchmark is employed.

    \textbf{LOL.} The LOL dataset has v1 and v2 versions. LOL-v2 is divided into real and synthetic subsets. The training and testing sets are split in proportion to 485:15, 689:100, and 900:100 on LOL-v1, LOL-v2-real, and LOL-v2-syn.

    \textbf{SID \& SMID.} For long-exposure evaluations, the Sony $\alpha$7S II subset of SID (2099/598 pairs) and the SMID dataset (15763/5046 pairs) are utilized, with RAW data converted to RGB using the standard in-camera signal processing (ISP) pipeline as defined in~\cite{SID}.

    \textbf{SDSD.} The static SDSD dataset, captured via a Canon EOS 6D Mark II with an ND filter, is adopted. It comprises 62/6 (indoor) and 116/10 (outdoor) pairs for training and testing.
    
    \textbf{Light100.} A standard split of 250 scenes is adopted (25,000 images) for training and 25 unseen scenes (2,500 images) for testing.

\subsubsection{Implementation Details}
Our framework is implemented in PyTorch and trained on NVIDIA GeForce RTX4090 GPUs. The AdamW optimizer is employed with $\beta_1 = 0.9$, $\beta_2 = 0.999$, and a weight decay of $10^{-4}$. The learning rate is initialized to $2 \times 10^{-4}$ and decayed over 1000 epochs with a cosine annealing schedule, following a 3-epoch warm-up. Training is performed on $3 \times 256 \times 256$ random crops with a batch size of 4, augmented with random flips and rotations. 

For evaluation, signal fidelity is assessed using Peak Signal-to-Noise Ratio (PSNR) and Structural Similarity (SSIM)~\cite{SSIM}. Additionally, Learned Perceptual Image Patch Similarity (LPIPS)~\cite{lpips} is adopted to quantify perceptual quality and texture realism. Model complexity is reported via GFLOPS (calculated on a $3 \times 256 \times 256$ input) and parameter count.

\begin{table*}[t]
\centering
\caption{Quantitative Comparison of  Standard LLIE Benchmarks on the LOL-v1~\cite{RetinexNet}, LOL-v2-Real~\cite{LOLv2}, LOL-v2-Syn~\cite{LOLv2}, SID~\cite{SID}, SMID~\cite{SMID}, SDSD-In~\cite{SDSD}, SDSD-Out~\cite{SDSD}. \textbf{Best} and \underline{second-best} results are highlighted.}
\label{tab1}
\scriptsize
\setlength{\tabcolsep}{2pt}
\begin{adjustbox}{width=\textwidth}
\begin{tabular}{@{}l cc cc cc cc cc cc cc cc@{}}
\toprule
\multirow{2}{*}{Methods} & \multicolumn{2}{c}{Complexity} & \multicolumn{2}{c}{LOL-v1} & \multicolumn{2}{c}{LOL-v2-Real} & \multicolumn{2}{c}{LOL-v2-Syn} & \multicolumn{2}{c}{SID} & \multicolumn{2}{c}{SMID} & \multicolumn{2}{c}{SDSD-in} & \multicolumn{2}{c}{SDSD-out} \\
\cmidrule(lr){2-3} \cmidrule(lr){4-5} \cmidrule(lr){6-7} \cmidrule(lr){8-9} \cmidrule(lr){10-11} \cmidrule(lr){12-13} \cmidrule(lr){14-15} \cmidrule(lr){16-17}
 & FLOPS (G) & Params (M) & PSNR & SSIM & PSNR & SSIM & PSNR & SSIM & PSNR & SSIM & PSNR & SSIM & PSNR & SSIM & PSNR & SSIM \\
\midrule
SID~\cite{SID} & 13.73 & 7.76 & 14.35 & 0.436 & 13.24 & 0.442 & 15.04 & 0.610 & 16.97 & 0.591 & 24.78 & 0.718 & 23.29 & 0.703 & 24.90 & 0.693 \\
DeepUPE~\cite{DeepUPE} & 21.10 & 1.02 & 14.38 & 0.446 & 13.27 & 0.452 & 15.08 & 0.623 & 17.01 & 0.604 & 23.91 & 0.690 & 21.70 & 0.662 & 21.94 & 0.698 \\
RetinexNet~\cite{RetinexNet} & 587.47 & 0.84 & 16.77 & 0.560 & 15.47 & 0.567 & 17.13 & 0.798 & 16.48 & 0.578 & 22.83 & 0.684 & 20.84 & 0.617 & 20.96 & 0.629 \\
EnGAN~\cite{EnGAN} & 61.01 & 114.35 & 17.48 & 0.650 & 18.23 & 0.617 & 16.57 & 0.734 & 17.23 & 0.543 & 22.62 & 0.674 & 20.02 & 0.604 & 20.10 & 0.616 \\
RUAS~\cite{RUAS} & 0.83 & 0.003 & 18.23 & 0.720 & 18.37 & 0.723 & 16.55 & 0.652 & 18.44 & 0.581 & 25.88 & 0.744 & 23.17 & 0.696 & 23.84 & 0.743 \\
FIDE~\cite{FIDE} & 28.51 & 8.62 & 18.27 & 0.665 & 16.85 & 0.678 & 15.20 & 0.612 & 18.34 & 0.578 & 24.42 & 0.692 & 22.41 & 0.659 & 22.20 & 0.629 \\
KinD~\cite{KinD} & 34.99 & 8.02 & 20.86 & 0.790 & 14.74 & 0.641 & 13.29 & 0.578 & 18.02 & 0.583 & 22.18 & 0.634 & 21.95 & 0.672 & 21.97 & 0.654 \\
SNR-Net~\cite{SNR-Net} & 26.35 & 4.01 & 24.61 & 0.842 & 21.48 & 0.849 & 24.14 & 0.928 & 22.87 & 0.625 & 28.49 & 0.805 & 29.44 & 0.894 & 28.66 & 0.866 \\
Retinexformer~\cite{Retinexformer} & 15.57 & 1.61 & \underline{25.16} & 0.845 & 22.80 & 0.840 & 25.67 & 0.930 & \textbf{24.44} & \textbf{0.680} & 29.15 & 0.815 & 29.77 & 0.896 & 29.84 & 0.877 \\
RetinexMamba~\cite{retinexmamba} & 34.75 & 4.59 & 24.03 & 0.827 & 22.45 & 0.843 & 25.88 & 0.933 & 22.45 & 0.656 & 28.62 & 0.809 & 28.44 & 0.894 & 28.52 & 0.859 \\
MambaLLIE~\cite{MambaLLIE} & 80.40 & 2.28 & 23.36 & 0.828 & 22.95 & 0.847 & 25.87 & 0.940 & 21.88 & 0.608 & 29.26& 0.818 & 30.12 & \underline{0.900} & 30.00 & 0.869 \\
URWKV~\cite{URWKV} & 18.34 & 2.25 & 24.22 & 0.855 & 23.11 & \textbf{0.874} & \underline{26.36} & \underline{0.944} & 23.11 & 0.673 & \underline{29.44} & \textbf{0.826} & \textbf{31.24} & \textbf{0.911} & 29.99 & \underline{0.887} \\
CIDNet~\cite{CIDNet} & 7.57 & 1.88 & 23.81 & \underline{0.857} & \textbf{24.11} & \underline{0.871} & 25.71 & 0.942 & 22.90 & \underline{0.676} & 27.37 & 0.785 & 29.02 & 0.877 & \underline{31.43} & 0.886 \\
CLE-RWKV (Ours) & 13.69 & 3.62 & \textbf{25.25} & \textbf{0.867} & \underline{23.30} & 0.861 & \textbf{26.75} & \textbf{0.947} & \underline{23.87} & 0.661 & \textbf{29.55} & \underline{0.820} & \underline{30.91} & 0.896 & \textbf{32.26} & \textbf{0.892} \\

\bottomrule
\end{tabular}
\end{adjustbox}
\end{table*}

\subsubsection{Performance on LLIE Benchmarks}
Prior to evaluating controllability, the fundamental enhancement capability of the proposed architecture is first assessed using a fixed-target variant, CLE-RWKV-Base. This serves to verify the backbone's proficiency in handling standard LLIE tasks when the illumination target is fixed, thereby isolating core performance from the influence of the control signal $\beta$. Such a configuration ensures that the model's ability to handle the complex coupling of underexposure and sensor noise is primarily attributable to the S2D architectural design rather than external guidance.

\begin{figure*}[!htbp]  
  \centering
  \includegraphics[width=1.0\linewidth]{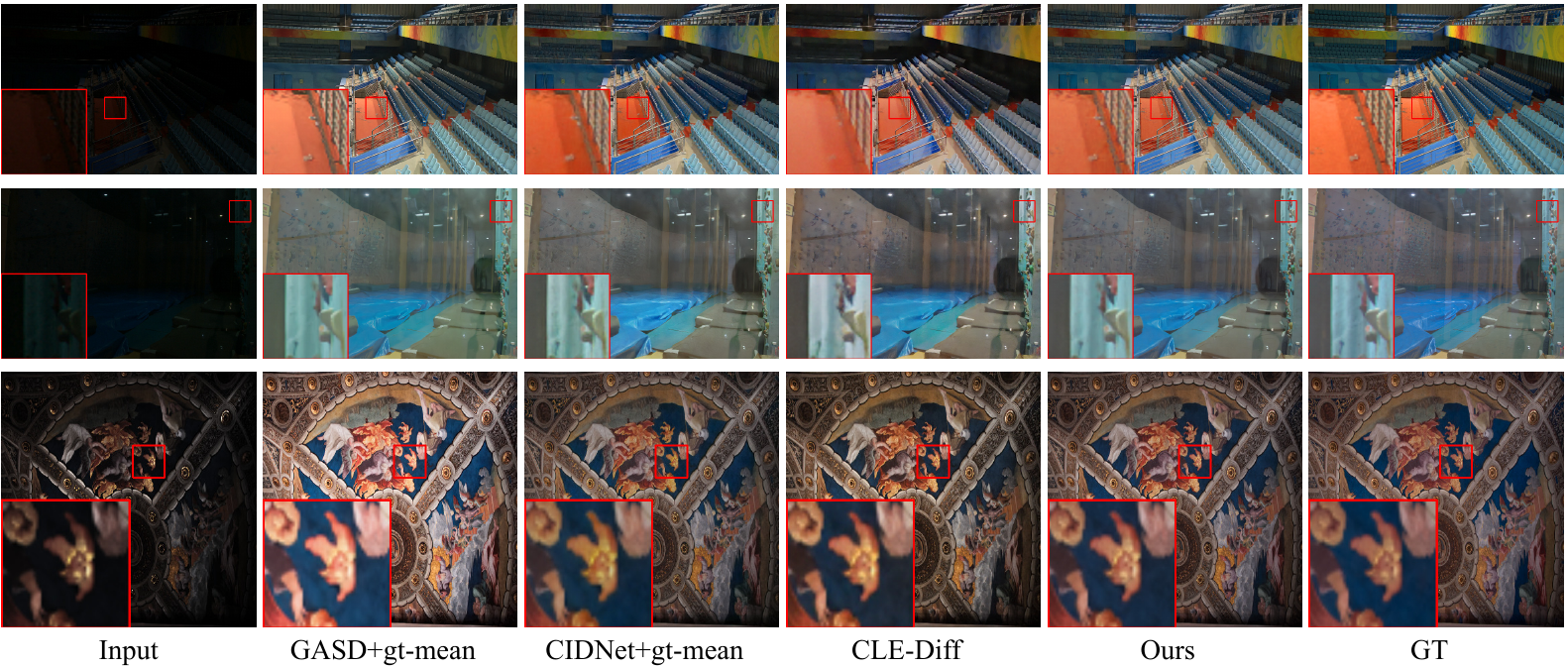}
  \caption{Visual comparisons on LOL-v1 (first row), LOL-v2-Real (second row), and LOL-v2-Syn (third row). The region within the red box is zoomed in and shown at the bottom left corner.}
  \label{fig:comp}
\end{figure*}

As summarized in Table \ref{tab1}, even without the control input, the base model ranks first on LOL-v1, LOL-v2-Syn, and SDSD-out. Specifically, it achieves a significant lead on LOL-v2-Syn and SDSD-out datasets. Visual comparisons in Fig.~\ref{fig:comp} show that our backbone recovers fine textures and balanced colors even in extremely dark regions. These results establish a state-of-the-art LLIE baseline, proving that the final controllable framework is built upon a robust and high-performance enhancement engine.

\subsubsection{Validation of the CLE Paradigm}
The core advantage of CLE is its ability to avoid the physical inaccuracies of post-hoc scaling (gt-mean). As analyzed in Sec. \ref{sec:theory}, simply scaling a model's output cannot correct for noise or ISP non-linearities.

Quantitative results in Table \ref{tab:cle_vs_gtmean} show that CLE-RWKV consistently outperforms the best baselines when tested on the same luminance targets. On LOL-v2-Real, our method achieves 29.85 dB PSNR, outperforming GSAD by 1.04 dB. The gap is even wider on Light100, where deterministic models struggle to adapt to 100 different light levels.

\begin{figure*}[!htbp]
  \centering
  \includegraphics[width=1.0\linewidth]{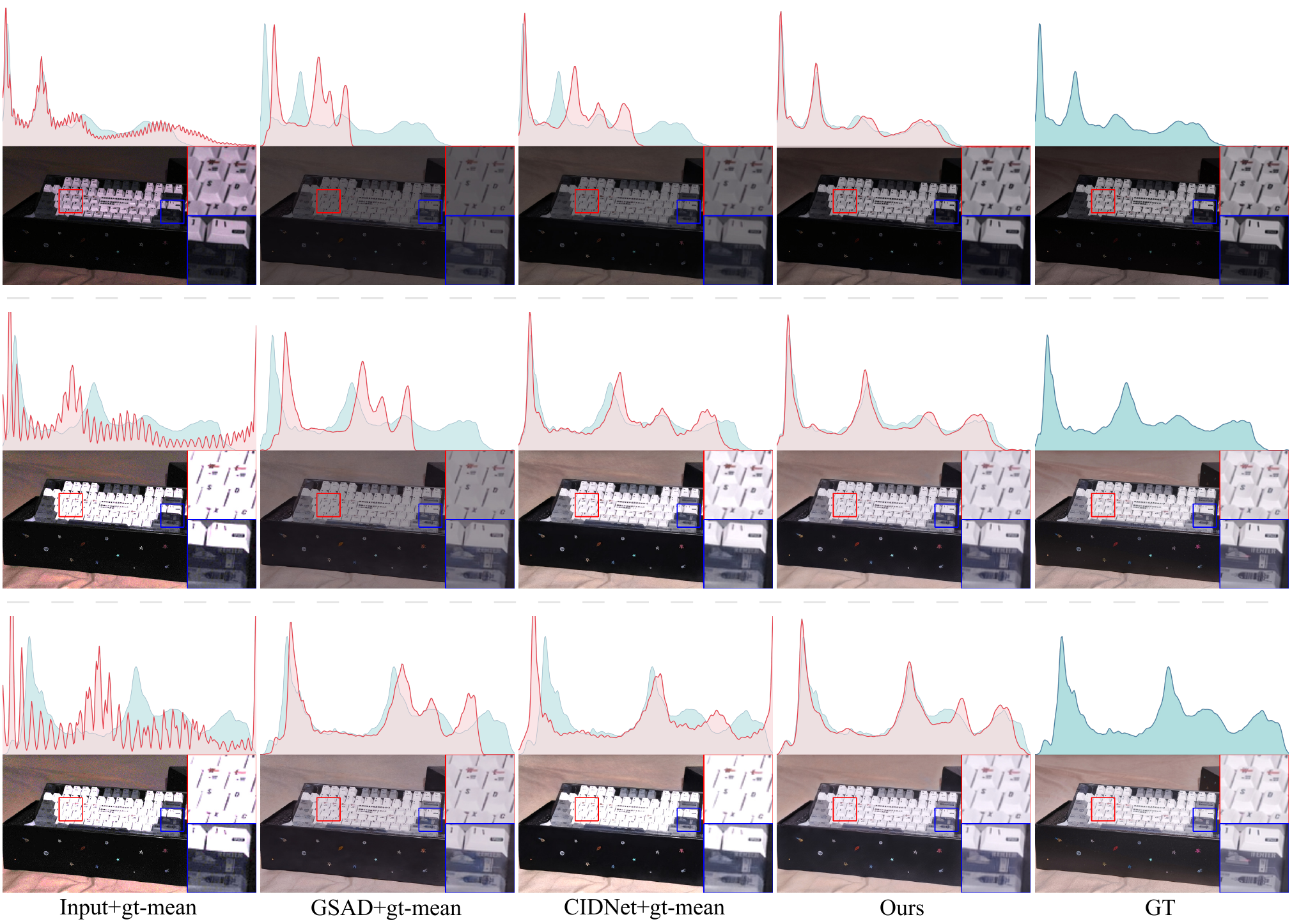}
  \caption{Images of the same scene at different luminance levels are presented in rows two, four, and six, with the corresponding luminance histograms (red) positioned above each image. For a clearer comparison, the GT histogram (blue) is overlaid onto the corresponding histograms in each row. While linear scaling tends to amplify noise and introduce distribution bias into deterministic baseline models, the proposed CLE-RWKV method achieves accurate histogram alignment and superior perceptual fidelity.
  }
  \label{fig:cle_comp}
\end{figure*}

The visual evidence in Fig.~\ref{fig:cle_comp} highlights these theoretical claims. In the first column, applying gt-mean to low-light inputs amplifies noise into jagged histogram spikes. In contrast, existing methods like GSAD~\cite{GSAD} and CIDNet~\cite{CIDNet} (middle columns) show a pervasive grayish cast and desaturated colors, even after mean correction. Their histograms fail to align with the ground truth, confirming the ``regression-to-mean" problem where models collapse toward a ``safe" but incorrect average brightness. Our CLE-RWKV (right column) produces realistic textures and accurate saturation, with histograms that closely match the ground truth.

\begin{table*}[t]
    \centering
    \caption{Quantitative Comparison of CLE Paradigm. We compare our method with state-of-the-art LLIE approaches on four benchmarks: LOL-v1~\cite{RetinexNet}, LOL-v2-Real~\cite{LOLv2}, LOL-v2-Syn~\cite{LOLv2}, and our Light100. Note that all baseline methods were retrained on the Light100 dataset for fair comparison. \textbf{Best} and \underline{second-best} results are highlighted.}
    \label{tab:cle_vs_gtmean}
    
    \small 
    \renewcommand{\arraystretch}{1.2} 
    \setlength{\tabcolsep}{0pt} 
    
    \begin{tabular*}{\textwidth}{l@{\extracolsep{\fill}}cccccccccccc}
    \toprule
    \multirow{2}{*}{\textbf{Method}} & \multicolumn{3}{c}{\textbf{LOL-v1}} & \multicolumn{3}{c}{\textbf{LOL-v2-Real}} & \multicolumn{3}{c}{\textbf{LOL-v2-Syn}} & \multicolumn{3}{c}{\textbf{Light100}} \\
    \cmidrule(lr){2-4} \cmidrule(lr){5-7} \cmidrule(lr){8-10} \cmidrule(lr){11-13}
     & PSNR & SSIM & LPIPS & PSNR & SSIM & LPIPS & PSNR & SSIM & LPIPS & PSNR & SSIM & LPIPS \\
    \midrule
    
    LLFlow+gt-mean~\cite{LLFlow} & 25.18 & 0.873 & 0.114 & 25.63 & 0.876 & 0.159 & 26.82 & 0.949 & 0.043 & 22.42 & 0.799 & 0.230 \\
    PyDiff+gt-mean~\cite{PyDiff} & 27.19 & 0.882 & 0.109 & 27.10 & 0.881 & 0.150 & 26.63 & 0.929 & 0.070 &  21.83 & 0.805 & 0.191 \\
    LLFormer+gt-mean~\cite{LLFormer} & 26.30 &  0.833 & 0.163 & 27.61 & 0.856 & 0.173 & 29.50 &  0.948 & \underline{0.045} & 25.71 & 0.863 & 0.121 \\
    GSAD+gt-mean~\cite{GSAD} & 27.84 & 0.877 &  0.091 & \underline{28.81} & \underline{0.895} & \underline{0.095} & 28.67 & 0.944 & 0.047 & 23.70 & 0.841 & 0.142 \\
    CIDNet+gt-mean~\cite{CIDNet} & \underline{28.20} & \underline{0.889} &  \underline{0.079} & 28.14 & 0.892 & 0.101 &  \underline{29.56} & \underline{0.950} & 0.047 & 25.48 &  \underline{0.869} &  0.087 \\
    
    \midrule
    
    CLE-Diff~\cite{CLE-Diff} & 25.51 & 0.825 & 0.164 & 26.84 & 0.829 & 0.207 & 29.10 & 0.932 & 0.050 & \underline{27.53} & 0.846 & \underline{0.063} \\

    CLE-RWKV (Ours)  & \textbf{29.18} & \textbf{0.892} & \textbf{0.062} & \textbf{29.85} & \textbf{0.902} & \textbf{0.088} & \textbf{34.05} & \textbf{0.968} & \textbf{0.021} & \textbf{29.21} & \textbf{0.903} & \textbf{0.056} \\
    \bottomrule
    \end{tabular*}
\end{table*}

\begin{figure}[t]
  \centering
  \includegraphics[width=1.0\linewidth]{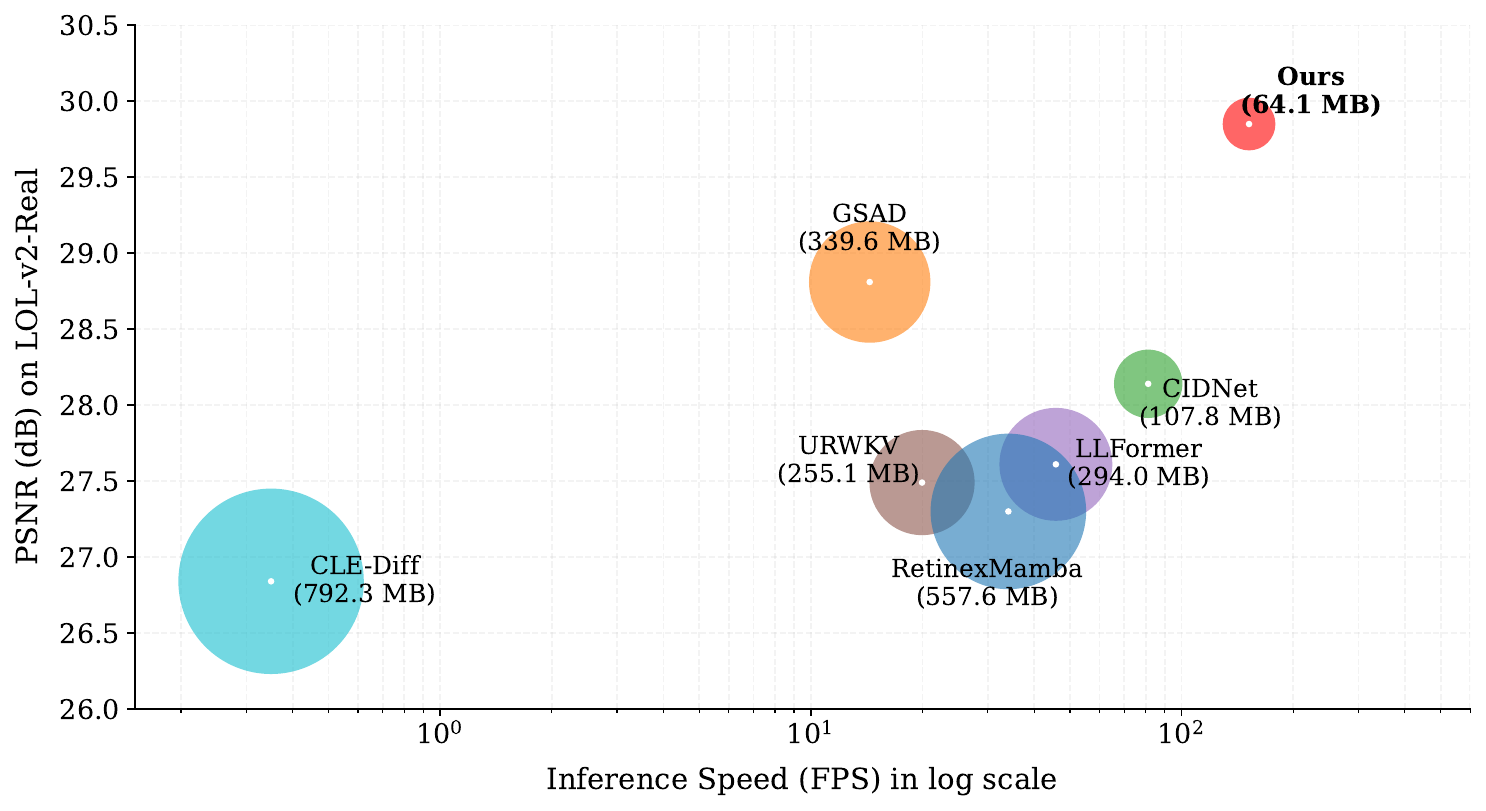}
  \caption{Efficiency Analysis on LOL-v2-Real. Each bubble represents a model, where the solid colored area corresponds to the peak memory footprint (labeled in MB) and the center white dot indicates the precise PSNR-FPS coordinate. All metrics are tested on an NVIDIA RTX 4090 with $3 \times 256 \times 256$ input. Our CLE-RWKV (red bubble) achieves the optimal trade-off across all metrics.}
  \label{fig:efficiency}
\end{figure}
 
\subsubsection{Efficiency Analysis}
The computational efficiency of the proposed CLE-RWKV is evaluated against existing models on the LOL-v2-Real dataset to determine their suitability for real-time applications. As illustrated in Fig.~\ref{fig:efficiency}, these methods are classified into three primary groups based on their underlying architectures: Diffusion-based, Transformer/CNN-based, and State Space Model (SSM)-based approaches.

The results in Fig.~\ref{fig:efficiency} show clear trade-offs between restoration quality and speed across these different categories. Diffusion-based models~\cite{CLE-Diff, GSAD} are capable of generating sharp high-frequency details but are hindered by their iterative sampling process. This recursive nature limits their performance to the sub-real-time range ($<$15 FPS) and leads to a heavy memory footprint. On the other hand, Transformer and CNN-based methods~\cite{CIDNet, LLFormer} provide a more balanced speed-to-quality ratio. However, these models still face challenges with either quadratic complexity or limited receptive fields, resulting in moderate frame rates and higher power consumption.

Models based on the SSM paradigm~\cite{URWKV, retinexmamba} generally show a much better efficiency profile due to their linear complexity. CLE-RWKV stands out even within this efficient group. While other SSM variants and Transformers typically cluster between 27 and 28.5 dB PSNR, our method reaches a peak performance of 29.85 dB on real-world test data. More importantly, by using the S2D design to optimize sequence length, CLE-RWKV maintains a high throughput of over 150 FPS while requiring only 64.07 MB of memory. These results indicate that CLE-RWKV is not only more effective at handling real-world noise and lighting challenges but is also the most practical choice for mobile devices where low-latency user interaction is a priority.

\begin{table}[t]
    \centering
    \caption{Ablation study. We evaluate each component's contribution. Speed (FPS) is measured on an RTX 4090 with $3 \times 256 \times 256$ input. $P$ denotes the patch size. \textbf{Best} and \underline{second-best} results are highlighted.}
    \label{tab:full_ablation}
    
    \footnotesize 
    \renewcommand{\arraystretch}{1.2} 
    \setlength{\tabcolsep}{4pt} 
    
    \begin{tabular}{llccccc}
    \toprule
    \multirow{2}{*}{Method} & \multirow{2}{*}{Configuration} & Speed & \multicolumn{2}{c}{LOL-v1} & \multicolumn{2}{c}{Light100} \\
    \cmidrule(lr){3-3} \cmidrule(lr){4-5} \cmidrule(lr){6-7}
     & & (fps) & PSNR & SSIM & PSNR & SSIM \\
    \midrule
    Color Space & sRGB & - & 27.06 & 0.871 & 29.07 & 0.902 \\
    \midrule
    \multirow{3}{*}{Architecture} 
    & UNet  & 33.4 & 27.19 & 0.872 & 28.59 & 0.899 \\
    & S2D ($P=1$) & 19.9 & 28.57 & 0.879 & 28.41 & 0.896 \\
    & S2D ($P=2$) &  \underline{60.4} & \underline{29.03} & \underline{0.890} & 28.97 & \underline{0.903} \\
    
    \midrule
    
    \multirow{2}{*}{Loss Func.} 
    & $\ell_1$ Only & - & 28.31 & 0.855 & \underline{29.18} & 0.900 \\
    & sRGB Only & - & 27.18 & 0.872 &  28.82 & 0.896 \\
    
    \midrule
    
    Proposed & Ours & \textbf{157.1} & \textbf{29.18} & \textbf{0.892} & \textbf{29.21} & \textbf{0.903} \\
    \bottomrule
    \end{tabular}
\end{table}

\subsection{Ablation Studies}
Detailed ablation experiments were conducted on LOL-v1 and Light100 to verify the effectiveness of our design choices. The results are summarized in Table \ref{tab:full_ablation}.

\textbf{Architecture Design.} The S2D design is first compared against a standard U-Net baseline using identical RWKV blocks. As shown in Table \ref{tab:full_ablation}, although the U-Net variant utilizes a higher computational budget ($\approx$17.53 GFLOPS), its throughput is limited to 33.4 FPS. This inefficiency stems from the $H \times W$ sequence length in the shallow layers, which creates a heavy burden for the linear scan. In contrast, CLE-RWKV reduces the sequence length immediately by folding spatial dimensions into channels. This architectural choice not only improves speed but also boosts PSNR from 27.19 dB to 29.18 dB on LOL-v1, proving that shorter, feature-rich sequences are more effective for SSM-based enhancement. Furthermore, the patch size $P$ is critical for addressing the ``scanning gap.'' While direct flattening ($P=1$) results in poor performance (28.57 dB) and low speed (19.9 FPS) due to the spatial distance between vertically adjacent pixels, increasing $P$ to 2 or 4 folds these neighbors into the channel dimension of a single token. The default $P=4$ configuration achieves the highest throughput of 157.1 FPS, confirming that larger patch sizes effectively recover the local inductive bias and bridge the scanning gap without needing auxiliary convolution branches.

\textbf{Color Space and Supervision.} The choice of color space and objective function is the final key factor. Training in the standard sRGB domain forces the model to amplify noise while boosting brightness, leading to a lower PSNR of 27.06 dB. Switching to the HVI space, which decouples intensity from chromaticity, yields a substantial +2.12 dB gain. Furthermore, experimental results indicate that using only $\ell_1$ loss leads to over-smoothed textures. By enforcing constraints in the HVI domain (Eq. \ref{eq:total_loss}) and including perceptual loss, the model achieves the best balance between signal fidelity (SSIM) and visual realism (LPIPS). This indicates that the proposed supervision strategy effectively helps the model reconstruct clean, noise-free textures while maintaining illumination accuracy.

\section{Conclusion}

In this paper, we propose CLE-RWKV, an efficient and controllable framework for real-world low-light image enhancement. The framework incorporates a Space-to-Depth (S2D) design to adapt State Space Models for dense restoration, combined with a noise-decoupled supervision strategy in the HVI color space. This approach effectively resolves the fundamental luminance mismatch in traditional deterministic models, eliminating the need for gt-mean post-processing. Moreover, we introduce Light100, the first real-world multi-illumination benchmark featuring 100 progressive lighting levels to enable precise and continuous controllability. In the future, we will delve deeper into the interplay between target luminance and complex ISP variables such as ISO and exposure time. Additionally, we will endeavor to extend the CLE paradigm to more generalized open-world environments, striving to provide a robust foundation for next-generation, user-centric enhancement systems.


\bibliographystyle{IEEEtran}

\bibliography{BibTeX}

\newpage

\vfill

\end{document}